# Verifier Theory and Unverifiability

**Roman V. Yampolskiy**
Computer Engineering and Computer Science
University of Louisville
roman.yampolskiy@louisville.edu

**Abstract**
Despite significant developments in Proof Theory, surprisingly little attention has been devoted to the concept of proof verifier. In particular, the mathematical community may be interested in studying different types of proof verifiers (people, programs, oracles, communities, superintelligences) as mathematical objects. Such an effort could reveal their properties, their powers and limitations (particularly in human mathematicians), minimum and maximum complexity, as well as self-verification and self-reference issues. We propose an initial classification system for verifiers and provide some rudimentary analysis of solved and open problems in this important domain. Our main contribution is a formal introduction of the notion of unverifiability, for which the paper could serve as a general citation in domains of theorem proving, as well as software and AI verification.

## 1. On Observers and Verifiers

The concept of an 'observer' shows up in contexts as diverse as physics (particularly quantum and relativity), biophysics, neuroscience, cognitive science, artificial intelligence, philosophy of consciousness, and cosmology [1], but what is an equivalent idea in mathematics? We believe it is the notion of the proof verifier. Consequently, the majority of open questions recently raised [1] by the Foundational Questions Institute related to the physics of the observer could be asked about proof verifiers. In particular, the mathematical community may be interested in studying different types of proof verifiers (people, programs, oracles, communities, superintelligences, etc.) as mathematical objects, ways they can be formalized, their power and limitations (particularly in human mathematicians), minimum and maximum complexity, as well as self-verification and self-reference in verifiers.

Proof Theory has been developed to study proofs as formal mathematical objects consisting of axioms from which, by rules of inference, one can arrive at theorems [2]. However, the indispensable concept of the verifier has been conspicuously absent from the discussion, particularly with regards to its formalization and practical manifestation. A *verifier* in the context of mathematics is an agent capable of checking a given proof, step-by-step, starting from axioms to make sure that all intermediate deductions are indeed warranted, that the final conclusion follows, and consequently, that the claimed theorem is indeed true. In this work we present an overview of different types of verifiers currently relied on by the mathematical community, as well as a few novel types of verifiers which we suggest be added to the repertoire of mathematicians at least as theoretical tools of *Verifier Theory*. Our general analysis should be equally applicable to different types of proofs (induction, contradiction, exhaustion, enumeration, refinement, nonconstructive, probabilistic, holographic, experiment, picture, etc.) and to computer software.

## 2. Historical Perspective

The field of mathematics progresses by proving theorems, which in turn serve as building blocks for future proofs of yet more interesting and useful theorems. To avoid introduction of costly errors in the form of incorrect theorems, proofs typically undergo an examination process, usually as a part of a peer-review. Traditionally, human mathematicians have been employed as proof verifiers; however, history is full of examples of undetected errors and important omissions even in the most widely examined proofs [3-7]. It has been estimated that at least a third of all mathematical publications contain errors [8]. To avoid errors and make the job of human verifiers as easy as possible "a single step in a deduction has been required … [t]o be simple enough, broadly speaking, to be apprehended as correct by a human being in a single intellectual act. No doubt this custom originated in the desire that each single step of a deduction should be indubitable, even though the deduction as a whole may consist of a long chain of such steps" [9].

Despite such stringent requirements, it has long been realized that a single human verifier is not reliable enough to ascertain validity of a proof with a sufficient degree of reliability. In fact, it is known that humans are subject to hundreds of well-known "bugs"[1], and probably many more unknown ones. To reduce the number of potential mistakes and to increase our confidence in the validity of a proof, a number of independent human mathematicians should examine an important mathematical claim. As Calude puts it "A theorem is a statement which could be checked individually by a mathematician and confirmed also individually by at least two or three other mathematicians, each of them working independently. But already we can observe the weakness of the criterion: how many mathematicians are to check individually and independently the status of [a conjecture] to give it a status of a theorem?" [4].

Clearly, the greater the number of independent verifiers, the higher is our confidence in the validity of a theorem. We can say that "a theorem is validated if it has been accepted by a general agreement of the mathematical community" [4]. Krantz agrees and says: "it is the mathematics profession, taken as a whole, that decides what is correct and valid, and also what is useful and is interesting and has value" [10]. Wittgenstein expresses similar views, as quoted in [11]: "who validates the 'mathematical knowledge'? … the acceptability ultimately comes from the collective opinion of the social group of people practising mathematics." So, for many practitioners of mathematics, proof verification is a social and democratic process in which "[a]fter enough internalization, enough transformation, enough generalization, enough use, and enough connection, the mathematical community eventually decides that the central concepts in the original theorem, now perhaps greatly changed, have an ultimate stability. If the various proofs feel right and the results are examined from enough angles, then the truth of the theorem is eventually considered to be established" [12].

While the mathematical community as a whole constitutes a powerful proof verifier, a desire for ever greater accuracy has led researchers to develop mechanized verification systems capable of handling formal proofs of great length. The prototype for such verifiers has its roots in *formal systems* [13] proposed by David Hilbert and which "contain an algorithm that mechanically checks the validity of all proofs that can be formulated in the system. The formal system consists of an alphabet of symbols in which all statements can be written; a grammar that specifies how the symbols are to be combined; a set of axioms, or principles accepted without proof; and rules of inference for deriving theorems from the axioms" [14]. However, there is a tradeoff when one switches from using human verifiers to utilizing automated ones, namely: "People are usually not very good in checking formal correctness of proofs, but they are quite good at detecting potential weaknesses or flaws in proofs" [15]. " 'Artificial' mathematicians are far less ingenious and subtle than human mathematicians, but they surpass their human counterparts by being infinitely more patient and diligent" [4]. In other words, while automated verifiers are excellent at spotting incorrect deductions, they are much worse than humans at seeing the "big picture" outlined in the proof.

---
[1] https://en.wikipedia.org/wiki/List_of_cognitive_biases

Additionally, to maintain a consistent standard of verification for all accepted theorems, a significant effort would need to be applied to reexamine already-accepted proofs. "to do so would certainly entail going back and rewriting from scratch all old mathematical papers whose results we depend on. It is also quite hard to come up with good technical choices for formal definitions that will be valid in the variety of ways that mathematicians want to use them and that will anticipate future extensions of mathematics. … [M]uch of our time would be spent with international standards commissions to establish uniform definitions and resolve huge controversies" [15].

Such criticism of automated verifiers is not new and has been expressed in the past, particularly from a human centric point of view: "No matter how precise the rules (logical and physical) are, we need human consciousness to apply the rules and to understand them and their consequences. Mathematics is a human activity" [4]. Additionally, "[m]echanical proof-checkers have indeed been developed, though their use is currently limited by the need or the proof to be written in precisely the right logical formalism" [16].

Despite such criticism, there is also a lot of hope in terms of what automated verification can offer mathematics. "[M]athematical knowledge is far too vast to be understood by one person, moreover, it has been estimated that the total amount of published mathematics doubles every ten to fifteen years… Perhaps computers can also help us to navigate, abstract and, hence, understand … proofs. Realising this dream of: computer access to a world repository of mathematical knowledge; visualising and understanding this knowledge; reusing and combining it to discover new knowledge" [17].

## 3. Classification of Verifiers

A certain connection exists between the concept of observer in physics and a verifier in mathematics/science. Both must be instantiated in the physical world as either hardware or software to perform its function, but other than that, we currently have a very limited understanding of types and properties associated with such agents. As the first step, we propose a simple classification system for verifiers, sorting them with respect to domain of application, type of implementation, and general properties. With respect to their domain, we see verifiers as necessary for checking mathematical proofs, scientific theories, software correctness, intelligent behavior safety, and consistency and properties of algorithms. Some examples:

- **Software Verifier** – evaluates correctness of a program. Via the Curry-Harvard Correspondence [18], proof verification and program verification are equivalent and software verification is a special case of theorem verification restricted to computational logic [19]. A compiler or interpreter can be seen as a program syntax verifier among other things.
- **AI-Verifier** – is a particular type of Software Verifier capable of verifying the behavior of intelligent systems in novel environments unknown at the time of design [20, 21]. Yampolskiy presents verification of self-improving software [22, 23] as a particular challenge to the AI community: "Ideally every generation of self-improving system should be able to produce a verifiable proof of its safety for external examination" [24]. Consequently, research linking functional specification to physical states is of great interest. "This type of theory would allow use of formal tools to anticipate and control behaviors of systems that approximate rational agents, alternate designs such as satisficing agents, and systems that cannot be easily described in the standard agent formalism (powerful prediction systems, theorem-provers, limited-purpose science or engineering systems, etc.). It may also be that such a theory could allow rigorously demonstrating that systems are constrained from taking certain kinds of actions or performing certain kinds of reasoning" [20].
- **Scientific Theory Verifier** – examines the output of computer simulations of scientific theories. A scientific theory cannot be considered fully accepted until it can be expressed as an algorithm and simulated on a computer. It should produce observations consistent with measurements obtained in the real world, perhaps adjusting for the relativity of time scale between simulation and the real world. In other words, an unsimulatable hypothesis should be considered significantly weaker than a simulatable one. It is possible that the theory cannot be simulated due to limits in our current computational capacity, hardware design, or capability of programmers and that it will become simulatable in the future, but until such time, it should have a tentative status. A scientific theory verifier could be seen as a formalized equivalent of a peer-reviewer in science.
- **NP Solution Verifier –** is an algorithm which can quickly (in polynomial time) check a certificate (also called witness) representing a solution, which can then be used to determine if a computation produces a "yes" or "no" answer. In fact, one of the requirements of NP-Completeness states that a problem is in that class if there exists a verifier for the problem. An NP-Completeness Verifier would check a reduction of a novel problem to an already known problem in the NP class to determine if it is of equal or lesser complexity. Analogously, we can postulate an AI-Completeness Verifier capable of checking if a problem is reducible to an instance of the Turing Test [25-27].

With respect to type, verifiers could be people (or groups of people), software, hypothetical agents such as oracles, or artificial (super)intelligent entities. For example:

- **Human Mathematician** – historically the default verifier for most mathematical proofs. Individual mathematicians have been recruited to examine mathematical reasoning since the inception of the field. Recent developments in computer-generated proofs appear to be beyond the capacity of human verifiers due to the size of such proofs.
- **Mathematical Community** – a collective of mathematicians taken as a whole used to examine and evaluate claimed proofs, while at the same time removing any outlier opinions of individual mathematicians. It is well known that the wisdom of crowds can outperform individual experts [28, 29].

- **Mechanical Verifier** (Automated Proof Checker) – automated software and hardware verifiers such as computer programs have been developed to assist in verification of formal proofs [30]. "The proof checker verifies that each inference step in the proof is a valid instance of one of the axioms and inference rules specified as part of the safety policy" [31]. They are believed to be more accurate than human mathematicians and are capable of verifying much longer proofs, which may not be surveyable [32-35] or too complex (not comprehensible [36]) for human mathematicians.
- **Hybrid Verifier** – a combination of other types of verifiers, most typically a human mathematician assisted by a mechanical verifier.
- **Oracle Verifier** – a verifier with access to an Oracle Turing Machine. Particular types would include a Halting Verifier (a hypothetical verifier not subject to the halting problem), a Gödel Verifier (not subject to incompleteness limits), and an undecidable proof verifier. All such verification would be done in constant time.
- **(Super)Intelligent Verifier** – a verifier capable of checking all decidable proofs, particularly those constructed by superintelligent AI.

Some verifiers also have non-trivial mathematical properties, which include: ability to self-verify, probabilistic proof checking, relative correctness, designated nature, meta-verification capacity, honest or dishonest behavior, and axiomatic acceptance. For example:

- **Axiomatically Correct Verifier** – a type of authority based verifier, which decides the truth of a theorem without a need to disclose its process. This is a verifier whose correctness is accepted without justification, much like an axiom is accepted by the math community.
- **Designated Verifier** – for some proofs of knowledge it is important that only the verifier nominated by the confirmer can get any confirmation of the correctness of the proof [37].
- **Honest (Trusted) Verifier** – "does not try to extract any secret from the prover by deviating from the proof protocol. … **Untrusted-Verifier** does not need to assume that the verifier is honest" [38].
- **Probabilistic Verifier** – a verifier, which by examining an ever-greater number of parts of a proof, arrives at a probabilistic measure of the correctness of the theorem. Such verifiers are a part of Zero Knowledge based protocols.
- **Relative Verifier** – a verifier with respect to which a particular theorem has been shown to be correct, which doesn't guarantee that it would be confirmed by other verifiers.
- **Gradual Verifier** – a verifier which determines a percentage of statements that are already guaranteed to be safe [39], permitting a gradual verification process to take place.
- **Meta-Verifier** – a hypothetical verifier capable of checking correctness of other verifiers.
- **Self-Verifier** – an agent which is capable of verifying its own accuracy [40]. A frequently suggested approach to avoid an infinite regress of verifiers, a self-verifying verifier could contain an error causing it to erroneously claim its own correctness [41] and is also subject to limitations imposed by Gödel's Incompleteness theorem [42] and other similar self-referential constraints [21].

## 4. Unverifiability

Unverifiability, an idea frequently discussed in philosophy [43-45], has been implicitly present in mathematics since the early days of the field. In this section, we survey literature that deals with the limits of proof verifiability caused by infinite regress of verifiers, and provides analysis of the concept of unverifiability. We believe that such explicit discussion will be useful to researchers interested in being able to cite this important idea, which so far has been relegated to the status of mathematical folklore [46] and only alluded to in the literature, despite being a more general result than incompleteness [42, 47].

*Unverifiability* is a fundamental limitation on verification of mathematical proofs, computer software, behavior of intelligent agents, and all formal systems. It is an ultimate limit to our ability to know certain information and is similar to other major "impossibilities" to acquiring knowledge in our universe such as: uncertainty [48], randomness [49, 50], incompleteness [42, 47], undecidability [51], undefinability [52], unprovability [53], incompressibility [14], noncomputability [54], and relativity [55]. Many paths can lead us to arrive at the concept of unverifiability, but in this paper we concentrate specifically on the infinite regress of verifiers.

For example, Calude et al. state: "what if the 'agent' human or computer checking a proof for correctness makes a mistake (agents are fallible)? Obviously, another agent has to check that the agent doing the checking did not make any mistakes. Some other agent will need to check that agent and so on. Eventually one runs out of agents who could check the proof and, in principle, they could all have made a mistake!" [56]. Later, Calude and Muller emphasize: "one cannot prove the correctness of the formal prover itself" [57]. Similarly, MacKenzie observes: "Indeed, if one was to apply the formal, mechanical notion of proof entirely stringently, might not the software of the automated theorem prover itself have to be verified formally? … The formal, mechanized notion of proof thus prompted a modern day version of Juvenal's ancient question, *quis custodiet ipsos custodes*, who will guard the guards themselves?" [58]. Others have expressed similar sentiments [11].

Our trust in a formal proof is only as strong as our trust in the verifier used to check the proof; as the verifier itself needs to be verified, and so on *ad infinitum*, we are never given a 100% guarantee of correctness, only asymptotically increasing probability of correctness. Worse yet, at the end of the chain of verifiers there is typically a single human, whose internal mechanism is simply not verifiable with our current technology and possibly not verifiable in principle. Additionally, problems other than infinite regress of verifiers may significantly reduce our ability to verify proofs. Such obstacles include: splicing and skipping [59], hidden lemmas [60], exponential size proofs [61] (with recent publication of a 200 terabyte computer proof [62] being only a current record which is unlikely to stand for long), impenetrable proofs [63], hardware failures [64, 65], Rice's theorem [66], and Gödel's Incompleteness theorem [42].

After the advent of probabilistic proofs by Rabin [67], "[s]ome have argued that there is no essential difference between such probabilistic proofs and the deterministic proofs of standard mathematical practice. Both are convincing arguments. Both are to be believed with a certain probability of error. In fact, many deterministic proofs, it is claimed, have a higher probability of error" [68]. "… the authenticity of a mathematical proof is not absolute, but only probabilistic. … Proofs cannot be too long, else their probabilities go down and they baffle the checking process.

To put it in another way: all really deep theorems are false (or at best unproved or unprovable). All true theorems are trivial" [3]. "A derivation of a theorem or a verification of a proof has only probabilistic validity. It makes no difference whether the instrument of derivation or verification is man or a machine. The probabilities may vary, but are roughly of the same order of magnitude" [3]. All proofs have a certain level of "proofness" [69], which can be made arbitrarily deep via expending necessary verification resources, but "in no domain of mathematics is the notion of provability a perfect substitute for the notion of truth [70]." To conclude, we reiterate Knuth's famous warning: "Beware of bugs in the above code: I have only proved it correct, not tried it."

## 5. Unverifiability of Software

Unverifiability has important consequences not just for mathematicians and philosophers of knowledge; more recently it has become an important issue in software and hardware verification, which can be seen as special cases of proof verification [18, 19]. Just like a large portion of published mathematical proofs, software is known to contain massive amounts of bugs [71], perhaps as many as fifty per thousand lines of code[2], but maybe as few as 2.3 [72]. Similarly, just like with mathematical proofs, the issue of infinite regress of verifiers is making software only probabilistically verifiable. For example, Fetzer writes: "There are no special difficulties so long as [higher-level machines'] intended interpretations are abstract machines. When their intended interpretations are target machines, then we encounter the problem of determining the reliability of the verifying programs themselves ("How do we verify the verifiers?"), which invites a regress of relative verifications" [73].

This notion of unverifiability of software has been a part of the field since its early days. Smith writes: "For fundamental reasons - reasons that anyone can understand - there are inherent limitations to what can be proven about computers and computer programs. … Just because a program is "proven correct" …, you cannot be sure that it will do what you intend" [74]. Rodd agrees and says: "Indeed, although it is now almost trite to say it, since the comprehensive testing of software is impossible, only very vague estimates of any program's reliability seem ever to be possible" [75]. Currently, most software is released without any attempt to formally verify it in the first place.

### 5.1 Unverifiability of Artificial Intelligence
One particular type of software, known as Artificial Intelligence (AI) (and even more so superintelligence), differs from other programs by its ability to learn new behaviors, adjust its performance, and act semi-autonomously in novel situations. Given the potential impact from intelligent software, it is not surprising that the ability to verify future intelligent behavior is one of the grand challenges of modern AI research [24, 76-78].

It has been observed that science frequently discovers so called "conjugate (complementary) pairs", "a couple of requirements, each of them being satisfied only at the expense of the other … . Famous prototypes of conjugate pairs are (position, momentum) discovered by W. Heisenberg in quantum mechanics and (consistency, completeness) discovered by K. Gödel in logic. But similar warnings come from other directions. According to Einstein …, 'in so far as the propositions of mathematics are certain, they do not refer to reality, and in so far as they refer to reality, they are not certain', hence (certainty, reality) is a conjugate pair" [56]. Similarly, in proofs we are "[t]aking rigour as something that can be acquired only at the expense of meaning and conversely, taking meaning as something that can be obtained only at the expense of rigour" [56]. With respect to intelligent agents, we can propose an additional conjugate pair - (capability, control). The more generally intelligent and capable an entity is, the less likely it is to be predictable, controllable, or verifiable.

---
[2] http://www.theengineer.co.uk/issues/may-2013-online/verification-system-aims-to-guarantee-software-function/

It is becoming obvious that just as we can only have probabilistic confidence in correctness of mathematical proofs and software implementations, our ability to verify intelligent agents is at best limited. As Klein puts it: "if you really want to build a system that can have truly unexpected behaviour, then by definition you cannot verify that it is safe, because you just don't know what it will do."[3] Muehlhauser writes: "The same reasoning applies to [Artificial General Intelligence] AGI 'friendliness.' Even if we discover (apparent) solutions to known open problems in Friendly AI research, this does not mean that we can ever build an AGI that is 'provably friendly' in the strongest sense, because … we can never be 100% certain that there are no errors in our formal reasoning. … Thus, the approaches sometimes called 'provable security,' 'provable safety,' and 'provable friendliness' should not be misunderstood as offering 100% guarantees of security, safety, and friendliness."[4] Jilk, writing on limits to verification and validation in AI, points out that "language of certainty" is unwarranted in reference to agentic behavior [79]. He also states: "there cannot be a general automated procedure for verifying that an agent absolutely conforms to any determinate set of rules of action."

Seshia et al., describing some of the challenges of creating Verified Artificial Intelligence, note: "It may be impossible even to precisely define the interface between the system and environment (i.e., to identify the variables/features of the environment that must be modeled), let alone to model all possible behaviors of the environment. Even if the interface is known, non-deterministic or over-approximate modeling is likely to produce too many spurious bug reports, rendering the verification process useless in practice. … [T]he complexity and heterogeneity of AI-based systems means that, in general, many decision problems underlying formal verification are likely to be undecidable. … To overcome this obstacle posed by computational complexity, one must … settle for incomplete or unsound formal verification methods" [78].

These results are not surprising. AI cannot be verified because AI itself can serve as a verifier which we already showed cannot be verified because of infinite regress problem and general unverifiability. By spending increasing computational resources, the best we can hope for is an increased statistical probability that our mathematical proofs, and software/AI are error free, but we should never forget that a 100% accurate verification is not possible, even in theory, and act accordingly. Artificial Intelligence, and even more so artificial Superintelligence, is unverifiable and so potentially unsafe [80-85].

## 6. Conclusions and Future Work
Our preliminary work suggests that "verifier" be investigated as a new mathematical object of interest for future study and opens the door for an improved understanding of the topic. For example, an artificially intelligent verifier could be used to re-check all previously published mathematical proofs, greatly increasing correctness of all proofs. Problems such as infinite regress of verifiers may be unsolvable, but methods such as probabilistic verification should be capable of giving us as much assurance as we are willing to pay for. Any progress in the proposed "verifier theory" will have additional benefits beyond its contribution to mathematics by making it possible to design safer advanced Artificial Intelligence, a topic that is predicted to become one of the greatest problems in science in the upcoming decades [86, 87]. A verifier is a hidden component of any proof; we can improve our capacity to verify by explicitly describing the required verification agent.

---
[3] https://intelligence.org/2014/02/11/gerwin-klein-on-formal-methods
[4] https://intelligence.org/2013/10/03/proofs/

It would be valuable to learn what types of physical or informational systems can act as verifiers and what their essential properties are. We should explore how selection of the type of the verifier influences mathematics as a field and specifically what categories of theorems we can prove and which we cannot prove with respect to different verifiers. Are there still undiscovered types of mathematical verifiers? Does a group of verifiers have greater power than the sum of its component modules? How can verifiers perform best while operating with limited computational resources? What is the formal relationship between the set of all verifiers and the set of all observers? Can a verifier be hacked and can the attack be contained in the proof it is examining? Can all these questions be reduced to a broader question on the nature of different possible types of intelligences [88]?

## Acknowledgements
The author is grateful to Kenneth Regan, Edward Frenkel, and Sebastien Zany for valuable advice on the topic. The author is particularly thankful to Yana Feygin, Søren Elverlin, Melissa Helton, David Kelley, and David Jilk for proofreading a draft of this work. Finally, I would like to emphasize that the argument presented in this paper is itself subject to unverifiability, and as such, is certainly not guaranteed to be correct, but probably is.